\pdfoutput=1

\documentclass[11pt]{article}

\usepackage[]{emnlp2021}

\usepackage{times}
\usepackage{latexsym}

\usepackage[T1]{fontenc}

\usepackage[utf8]{inputenc}

\usepackage{microtype}
\usepackage{amssymb}
\usepackage{amsmath}
\usepackage{graphicx}

\title{FLiText: A Faster and Lighter Semi-Supervised Text Classification with Convolution Networks}

\author{Chen Liu, Mengchao Zhang, Zhibin Fu, Pan Hou, Yu Li\thanks{~ Corresponding author.}\\
        Beijing Value Simplex Technology Co. Ltd \\ \texttt{liuchen1350@gmail.com} \\ \texttt{liyu@entropyreduce.com}}

\begin{document}
\maketitle
\begin{abstract}
In natural language processing (NLP), state-of-the-art (SOTA) semi-supervised learning (SSL) frameworks have shown great performance on deep pre-trained language models such as BERT, and are expected to significantly reduce the demand for manual labeling. However, our empirical studies indicate that these frameworks are not suitable for lightweight models such as TextCNN, LSTM and etc. In this work, we develop a new SSL framework called FLiText, which stands for Faster and Lighter semi-supervised Text classification. FLiText introduces an inspirer network together with the consistency regularization framework, which leverages a generalized regular constraint on the lightweight models for efficient SSL. As a result, FLiText obtains new SOTA performance for lightweight models across multiple SSL benchmarks on text classification. Compared with existing SOTA SSL methods on TextCNN, FLiText improves the accuracy of lightweight model TextCNN from 51.00\% to 90.49\% on IMDb, 39.8\% to 58.06\% on Yelp-5, and from 55.3\% to 65.08\% on Yahoo. In addition, compared with the fully supervised method on the full dataset, FLiText just uses less than 1\% of labeled data to improve the accuracy by 6.59\%, 3.94\%, and 3.22\% on the datasets of IMDb, Yelp-5, and Yahoo respectively.
\end{abstract}

\section{Introduction}

Developments in deep learning technology have great breakthroughs in most natural language processing (NLP) tasks, such as machine translation, sentiment analysis, and reading comprehension.~\citep{BERT,xlnet,flat,span-re,Summarization,chatbot,MT,qa,skep,CWS}
The success of these advancements is highly dependent on large-scale and high-quality manual labeled data. However, obtaining vast amounts of high-quality labeled data is expensive. Especially in certain fields, such as finance, medicine, law, and so on, text labeling relies on the in-depth participation of field experts. The rapid development of SSL technology is expected to significantly reduce the demand for labeled data. The core goal of this technology is to use a small number of labeled data and vast amounts of unlabeled data to train a model with good generalization performance to solve machine-learning problems.~\citep{PseudoLabel,temporal,virtual,mean-teacher,mixmatch,uda,fixmatch,remixmatch}
\begin{figure}[ht]
\centering
\includegraphics[scale=0.5]{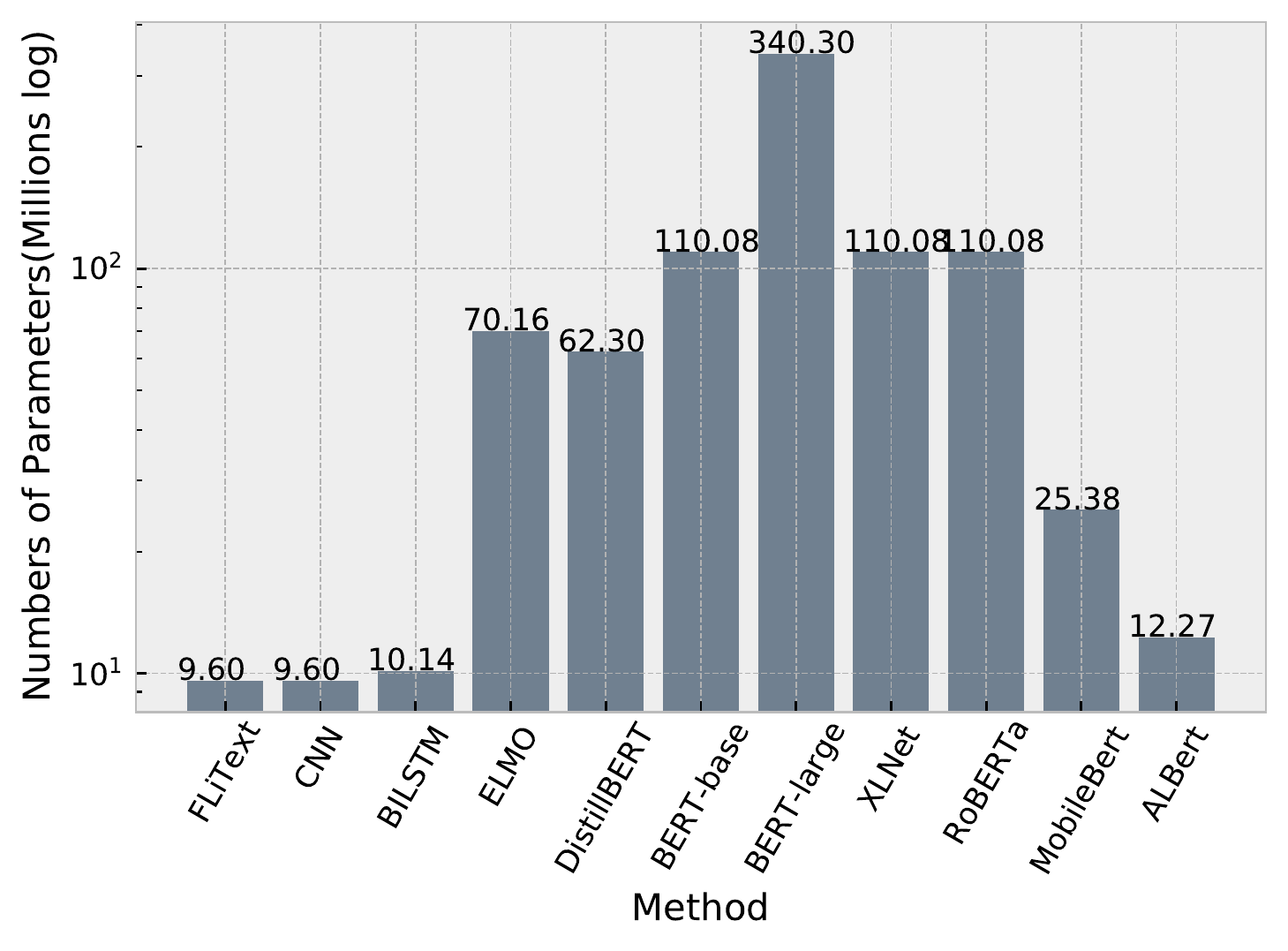}
\caption{Comparison of the scale of parameters between FLiText and pre-trained language models in recent years.}
\label{fig:pathdemo4}
\end{figure}
Unsupervised data augmentation (UDA)~\citep{uda} and  MixText~\citep{mixtext} are SOTA SSL methods for text classification, and have been used to various tasks with notable success. In NLP, applying the SSL framework to deep pre-trained language models (e.g., BERT, GPT, and XLNet) has been demonstrated effective. However, the good performance of these SSL methods depends on a bulky ``large model''. In most practical situations, due to the large-scale parameters and slow inference speed, it is difficult to implement these models with limited resources, such as mobile devices~\citep{mobilebert}, online search engines~\citep{twinbert}, and edge platforms~\citep{edgebert}. 
An intuitive idea to address the problem is to apply the SSL method on a small model, such as TextCNN, of which the parameter scale is about one or two orders of magnitude lower than that of BERT, as shown in Figure 1. However, many applications show that the existing SOTA SSL framework performs poorly on lightweight models. Furthermore, there is a lack of relevant research on the implementation of SSL on lightweight models.

This paper develops an SSL framework on lightweight models, for faster and lighter semi-supervised text classification (FLiText). We use a deep pre-trained inspirer network to learn the distribution relationship and the task-specific features of the data. Next, the inspirer network provides two types of regularization constraints on a lightweight model. The intuitive explanation is as follows: ``Teachers not only teach results but also teach experiences in the learning process so that students can learn more effectively.''
To evaluate FLiText, we compare FLiText and SOTA methods on three benchmark text classification datasets. We also conduct an ablation study to verify the performance of each part of FLiText. The results show that FLiText can significantly improve the inference speed while maintaining or exceeding SOTA performance. Compared with UDA on TextCNN, FLiText improves the accuracy from 51.00\% to 90.32\% on the IMDb dataset, from 39.80\% to 58.06\% on the Yelp-5 dataset, and from 55.30\% to 65.08\% on the Yahoo dataset. Compared with the supervised learning on complete datasets, the performance is improved by 6.28\%, 4.08\%, and 3.81\% on the three datasets respectively, by just using less than 1\% labeled data. Our contributions can be summarized as follows:
\begin{itemize}
\item To our best of knowledge, in NLP, FLiText is the first SSL framework proposed for lightweight models, which can achieve new SOTA SSL performance on multiple datasets.
\item We experimentally demonstrate that FLiText using less than 1\% labeled data outperforms the supervised method using complete datasets on a lightweight model.
\item We propose a new semi-supervised distillation method for knowledge distilling from BERT to TextCNN, which outperforms output-based knowledge distillation (KD) significantly.
\item We experimentally demonstrate introducing a consistent regularization framework in KD improves the performance of the student model. Our source code can be obtained from: \url{https://github.com/valuesimplex/FLiText}.
\end{itemize}

\section{Related Work}
\paragraph{Semi-Supervised Learning:}~\citep{PseudoLabel} uses the pseudo labels and the unlabeled data for supervised learning.
~\citep{Ladder} obtains the learning signal by autoencoder.
~\citep{temporal} calculates the mean square error between the prediction of the current model and the average of the historical prediction to construct the consistency regularization.
~\citep{virtual} adopts the method of adversarial learning to generate noise.
~\citep{mixmatch} uses the average of the prediction of K types of data augmentation on unlabeled data to achieve consistency regularization. 
~\citep{remixmatch} aligns the predicted distribution with the ground-truth distribution.
UDA~\citep{uda} achieves consistency regularization on unlabeled data after back translation and tf-idf representation.
~\citep{mixtext} proposes ``TMix'' data augmentation.
~\citep{ren2020unlabeled} adds weights for each unlabeled sample.
 However, these SOTA methods all rely on the deep pre-trained language model such as BERT, and so far no research on the SSL on lightweight models has been shown.
\paragraph{Knowledge Distillation:}~\citep{hinton} uses student model to mimic teacher's prediction by soft target. 
~\citep{distilllstm} distills BERT into a single layer of BILSTM. For the first time, the knowledge of the Transformer-based model was distilled into the non Transformer-based model.
~\citep{BERT-PKD} extracts knowledge from the intermediate layer of the BERT; 
~\citep{DistilBERTAD} distills knowledge during the model’s pre-trained stage;
~\citep{TinyBERT} combines the above various methods and propose a two-stage distillation method.
Although all of these methods have achieved excellent results, the transformer has the problem of a huge amount of parameters and high computational complexity.

\section{Method}
\subsection{Framework}
As shown in Figure 2, the biggest difference between FLiText and the previous SSL model is the introduction of an inspirer network outside the lightweight target network. The inspirer network utilizes consistency regularization and data augmentation technology to sufficiently mine information and features from the unlabeled data and limited labeled data. Then it provides a regularized constraint on two levels (i.e., output and hidden spaces) to lead the lightweight target network to realize efficient SSL using only a few labeled data.
\begin{figure*}[ht]
\centering
\includegraphics[scale=0.7]{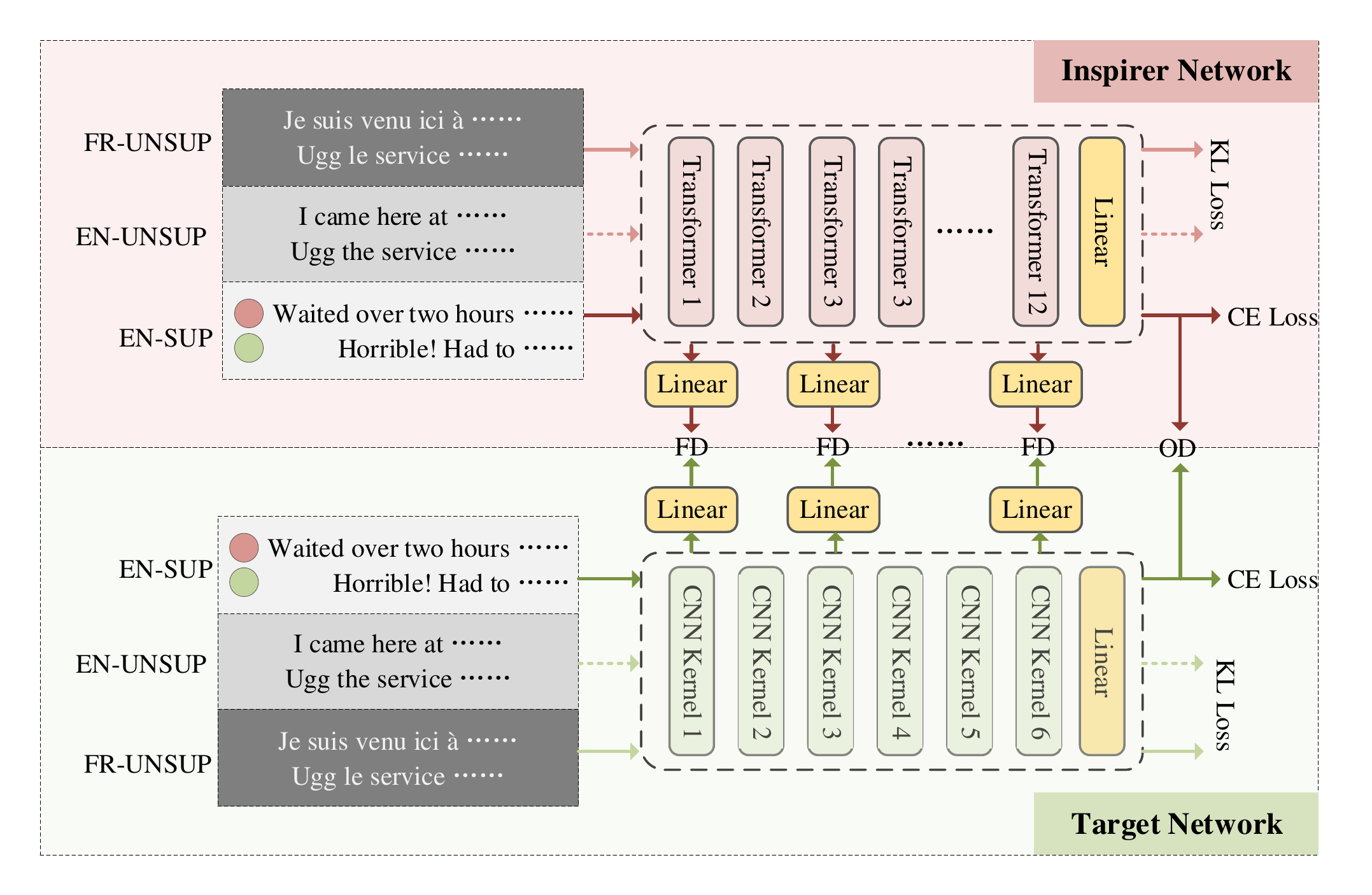}
\caption{The architecture of FLiText. EN-SUP is labeled data of which the language is English; The red and green circles denote the different category of the text; EN-UNSU is unlabeled data of which the language is English and FR-UNSUP is noise version of the unlabeled data of which the language is French; FD represents the feature-based distillation loss designed by FLiText; OD represents the output-based distillation loss.}
\label{fig:pathdemo4}
\end{figure*}
The entire framework comes from two types of insights. First, ~\citep{deepneed} mentions that training a lightweight model based on the output of a larger model would be better than on the original data. Additionally, the general approximation theorem~\citep{UAT} identifies that most functional spaces contained in lightweight models could cover the target function required by the downstream task. Therefore, as a supplement to the current optimizer, the inspirer network can provide a well-qualified regularized constraint for the training of the lightweight model.

We define  $X={(x_i,y_i),i\in(1,...,n)}$ as the labeled dataset, $U={(u_j),j\in(1,...,m)}$ as the unlabeled dataset, where 
$n$ is the number of labeled samples,
$m$ is the number of unlabeled samples.

\subsubsection{Inspirer Network}
The inspirer network comprises three parts: text encoder, text classifier, and feature projection. The text encoder is a pre-trained language model stacked with multiple transformer structures, such as BERT. Given an input sentence $x_i$, we can obtain a representation of the feature vector of ``[CLS]'' from BERT, where $h_i\in{\mathbb{R}^d}$

\begin{equation}\label{1}
h_i=BERT(x_i)
\end{equation}
where $d$ is the dimension of hidden vector.  

We use $h_i$ and a two-layer multi-layer perceptron (MLP) to construct the text classifier and fine-tuning the downstream classification task. We denote the result obtained by the MLP as $z_i^{(T)}$:

\begin{equation}\label{2}
z_i^{(T)}=MLP(h_i)
\end{equation}

To align the dimensions of BERT and TextCNN, we feed hidden state into the feature projection, $Ig(\cdot)$, which compose of a single MLP and a nonlinear activation function. The output can be denoted as $If_i^l$, where $l\in{L}$ represents the number of transformer layers.

\vspace{-2ex}
\begin{equation}\label{3}
Ig(\cdot)=Tanh(MLP(\cdot))
\end{equation}
\vspace{-2ex}
\begin{equation}\label{4}
If_i^l=Ig(Transformer(x_i))
\end{equation}
\subsubsection{Target Network}
The target network comprises a text encoder, a text classifier, and a feature projection. We use TextCNN~\citep{textcnn} as the text encoder. 

Because of its lightweight and parallelism, it has been broadly applied to all types of text-treatment systems~\citep{chatbot,MT,qa,skep,CWS}. Given an input sentence $x_i$, we use TextCNN to extract its information and the max-pooling operation to obtain its vector representation, $c_i\in{\mathbb{R}^d}$

\begin{equation}\label{5}
c_i=MaxPool(CNN(x_i))
\end{equation}
where $d$ represents the dimension of hidden vector output by TextCNN.

We use $c_i$ and an MLP to construct a text classifier for the downstream text classification task. We denote the result obtained by the MLP as $z_i^{(s)}$
\begin{equation}\label{6}
z_i^{(s)}=MLP(c_i)
\end{equation}

The structure of the feature projection is the same as that of the inspirer. The difference is that we use the feature map to replace the output of the transformer layer in the inspirer:

\vspace{-2ex}
\begin{equation}\label{7}
Tg(\cdot)=Tanh(MLP(\cdot))
\end{equation}
\vspace{-2ex}
\begin{equation}\label{8}
Tf_i^k=Tg(CNN(x_i))
\end{equation}
$Tg(\cdot)$ is the feature  projection of the target network, and $Tf_i^k$ is the projection representations of $th_k$ CNN filter.

\subsection{Two-stage Learning}
FLiText consists of two training stages: Inspirer pre-training and Target network training. In the first stage, we introduce a variety of advanced semi-supervised ideas to complete the inspirer’s training at downstream tasks. During the second stage, FLiText maintains the inspirer’s parameters unchanged and guides the training of the target network in the downstream tasks via multi-level regular constraints provided by the inspirer network, ultimately achieving efficient semi-supervised distillation learning. By means of the two-stage training operation, FLiText finally completes the SSL on the lightweight target network.

\subsubsection{Inspirer Network Training}
The training method is inspired by a consistency regularization framework. The loss function consists of two parts: the cross-entropy loss applied to labeled data and the consistent regularization loss on the unlabeled data.
 Similar to ~\citep{uda}, to restrain over-fitting, we also use training-signal annealing technology to balance the participation of labeled data in the training process. Given unlabeled data $u_i$ and its noise version $a_i\in{X^a}$, we calculate the inspirer training loss:

\begin{small}
\begin{equation}\label{9}
L_{CE}=\sum_{i\in{N}}\sum_{c\in{C}}y,log(p(x_i,\theta)))
\end{equation}
\end{small}
\vspace{-2ex}
\begin{small}
\begin{equation}\label{10}
L^{(T)}=L_{CE}+KL(p(u_i,\theta)^{(T)},p(a_i,\theta)^{(T)})
\end{equation}
\end{small}
where the superscript $u$ is the unlabeled data identifier, $a$ is the noise data identifier, $(T)$ is the inspirer identifier, $N$ is the number of labeled samples, and $C$ is the total number of labeled categories. $p(\cdot)$ is the predicted probability distribution produced by the model for input $x$ and parameter $\theta$. $L_{CE}$ is the standard cross-entropy loss applied to labeled samples and $L^{(T)}$ is the objective function of the inspirer. Other symbols are the same as before.

\subsubsection{Target Network Distillation}
In FLiText, we use two types of distillation methods together with the consistency regularization framework to complete the guidance of the inspirer network to the target network, by applying a regularized constraint to the objective function of the target network.
\paragraph{Output-based Distillation.} Like~\citep{mukherjee} , we also use hard label or soft label for output-based KD method:

\begin{equation}\label{11}
L_{soft}=\left \|z_i^{(T)}-z_i^{(S)} \right \|_2^2
\end{equation}




\begin{equation}\label{12}
y_i^{T}=argmax(p(x_i,\theta)^{(T)})
\end{equation}
\begin{equation}\label{13}
L_{hard}=CE(p(x_i,\theta)^{(S)},y_i^{(T)})
\end{equation}
where $y_i^{(T)}$ is the predicted label.

\paragraph{Feature-based Distillation.} Due to the output-based KD method does not account for the intermediate learning process, we next introduce another KD method: feature-based KD. ~\citep{bertlayers} shows that BERT can capture surface, syntactic and semantic representations from low-level layer to high-level layer. Inspired by this, considering that the text features extracted by the CNN filters with different sizes are different, FLiText assumes that \textbf{the linguistic level of features captured by the CNN filters increases with their size.} For example, a convolution having a window size of 4 is mainly focus on word-level features, whereas filters having a window size of 15 can capture semantic-level features. As shown in Figure 3, the proposed hidden-space feature-based distillation scheme can achieve knowledge transfer from BERT to TextCNN. In this scheme, we align the small-size filters with the lower layers of the BERT and the large-size ones with the higher layers. This is equivalent to imposing an a priori constraint on TextCNN. Namely, small filters are required to capture word-level features, medium filters capture syntactic features, and large ones capture semantic features.
\begin{figure}[ht]
\centering
\includegraphics[scale=0.4]{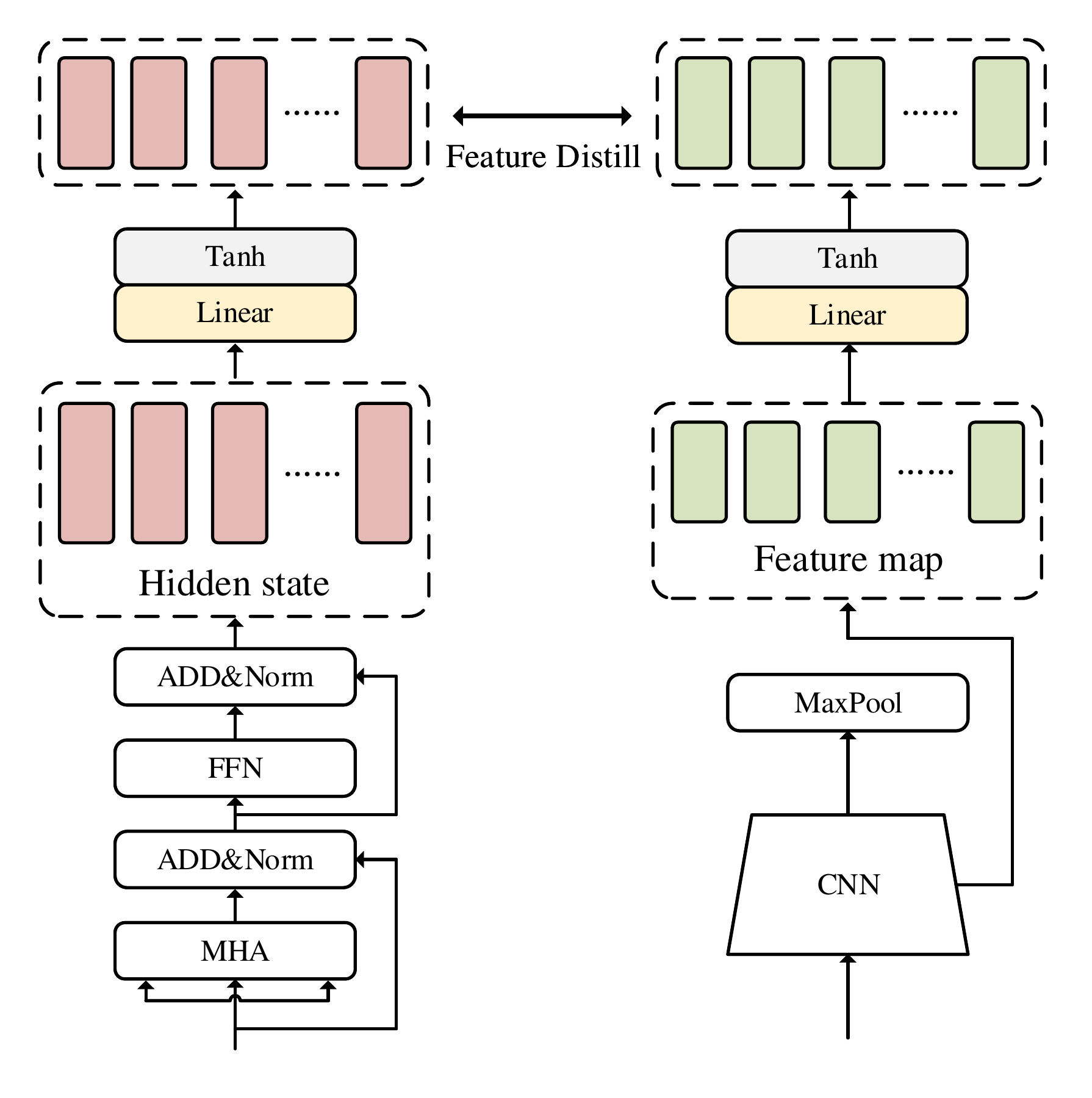}
\caption{Architecture of Feature Distillation.}
\label{fig:pathdemo4}
\end{figure}
We use the feature projection to match the transformer's hidden states and feature maps. We complete the knowledge extraction by minimizing the mean-squared error between the two feature projections, which is recorded as the feature distillation loss, $L_{feature\_distill}$.

\vspace{-1ex}
\begin{equation}\label{14}
L_{feature\_distill}=MSE(If_i^l,Tf_i^k)
\end{equation}

\paragraph{Consistency Regularization.}Owing to the differences in the parameter space and the network structure between the target and inspirer networks, there is a problem of knowledge loss during the learning process. If only the KD methods are adopted, the target network would not be able to learn some of the functional characteristics of the inspirer network. Therefore, we introduce consistency regularization to constrain the target network, which keeps it smooth enough in the function space. Thus, the network should be flat around the input data. Even if the input data change slightly or their forms change while remaining semantically unchanged, the output of the model can remain basically the same. This is consistent with the training of the inspirer network:

\begin{small}
\begin{equation}\label{15}
L_{consist}^{(S)}=KL(p(u_i,\theta)^{(S)},p(a_i,\theta)^{(S)})
\end{equation}
\end{small}
Finally, the loss function of the target network is

\begin{small}
\begin{equation}
\begin{aligned}
\label{16}
L_{total}=L_{CE}+L_{distill}^{(sup)}+L_{feature\_distill}^{(sup)}\\ 
+L_{distill}^{(unsup)}+L_{feature\_distill}^{(unsup)}+L_{consist}^{(S)}
\end{aligned}
\end{equation}
\end{small}
where the superscript $(sup)$ is the labeled sample identifier, $(unsup)$ is the unlabled sample identifier. $L_{CE}$ represents the classification loss calculated on labeled data. $L_{distill}^{(sup)}$ is the output-based distillation loss on the labeled data; $L_{feature\_distill}^{(sup)}$ is the feature-based distillation loss on the labeled data; and $L_{consist}^{(S)}$ is the consistency regularity loss of TextCNN.

\section{Experiments}
\subsection{Dataset}
We verify the performance of FLiText on three publicly available English text classification benchmark datasets: IMDb~\citep{IMDb}, Yahoo~\citep{yahoo} and Yelp-5~\citep{uda}. 
From Yahoo and Yelp-5, we randomly sample 70,000 sentences of unlabeled data, and 5,000 sentences as test data to verify the SSL method. We also randomly select 70,000 sentences of labeled data as a full dataset for the supervision method. For all datasets, we use French as an intermediate language for back translation. Table 1 shows the statistical information.


\subsection{Implementation Details}
In all experiments, we set the max sentence length to be 256.
The dropout rate is 0.5. We use Adam to optimize the parameters of each model. 
It is found that all of the methods including the proposed work in this paper and other methods for comparison can achieve the best performance within 10 epochs. In order to ensure the consistency of the experimental conditions, 10 epochs are uniformly used.
 For the inspirer network, we use BERT-based-uncased\footnote{\url{https://github.com/google-research/bert}} as the encoder, a two-layer MLP with 768 hidden states, and tanh as the activation function. The learning rate is 2e-5 for the BERT encoder and 1e-3 for the MLP model. For the target network, we use the Glove\footnote{\url{https://apache-mxnet.s3.cn-north-1.amazonaws.com.cn/gluon/embeddings/glove/glove.6B.zip}} 300d vector as the embedding layer initialization parameter, TextCNN as the encoder. We use filters with sizes of 2, 3, 5, 7, 9 and 11 respectively. The number of output channels is 200, and the max-pooling operation is used to extract key information. For the project layer, we use a single layer MLP with a hidden size of 256 and a Relu as the activation function. Most of the reports ~\citep{mixtext,uda,fixmatch,remixmatch} about SSL only report accuracy or error rate. Hence, we also use accuracy for comparison with other works in this paper.
\begin{small}
\begin{table}
\centering
\begin{tabular}{lccccc}
\hline
\textbf{Dataset} & \textbf{Class} & \textbf{Labeled} & \textbf{Dev} & \textbf{Test} \\
\hline
IMDb & 2 & 25000 & 25000 & 25000 \\
Yelp-5 & 5 & 70000 & 5000 & 5000 \\
Yahoo & 10 & 70000 & 5000 & 5000 \\
\hline
\end{tabular}
\caption{Statistics of the IMDb, Yahoo and Yelp-5.}\label{tab:accents}
\end{table}
\end{small}
\subsection{Result}
\begin{small}
\begin{table*}
\centering
\begin{tabular}{l|ccc|ccc|ccc}
\hline
\textbf{Model} & \multicolumn{3}{c|}{\textbf{IMDb}} & \multicolumn{3}{c|}{\textbf{Yelp-5}} & \multicolumn{3}{c}{\textbf{Yahoo}} \\
\cline{2-10}
& 20 & 500 & 2500 & 500 & 1000 & 2500 & 500 & 1000 & 2500 \\
\cline{1-10}
TextCNN(fully) & \multicolumn{3}{c|}{83.9} & \multicolumn{3}{c|}{54.12} & \multicolumn{3}{c}{61.86} \\
\cline{2-10}
UDA(BERT) & 90.15 & 90.27 & \textbf{91.07} & 56.53 & \textbf{59.64} & \textbf{61.37} & 66.86 & \textbf{68.9} & 70.32 \\
MixText & 78.24 & 88.17 & 90.02 & 54.34 & 57.98 & 60.02 & \textbf{67.38} & 68.84 & \textbf{70.4} \\
UDA(TextCNN) & 51.00 & 78.38 & 80.74 & 39.8 & 42.6 & 48.12 & 55.3 & 55.58 & 62.32 \\
TextCNN(KD) & 90.32 & 90.43 & 90.21 & 56.87 & 57.98 & 58.35 & 64.23 & 65.81 & 66.75 \\
UDA(ALBERT) & 88.24 & 89.04 & 90.07 & 51.08 & 55.43 & 57.44 & 62.13 & 64.81 & 66.07\\
UDA+KD(DistilBERT$_\text{6}$) & 89.51 & 90.42 & 91.06 & 57.21 & 58.41 & 59.87 & 55.09 & 60.64 & 66.97 \\
UDA+KD(TinyBERT$_\text{4}$) & 87.17 & 88.23 & 89.17 & 56.51 & 57.79 & 58.97 & 64.67 & 66.17 & 67.46 \\
UDA+KD(TinyBERT$_\text{6}$) & 87.54 & 89.26 & 90.34 & 56.41 & 57.94 & 59.43 & 66.01 & 67.87 & 69.76 \\
FLiText & \textbf{90.49} & \textbf{90.74} & 90.96 & \textbf{58.06} & 59.27 & 60.4 & 65.08 & 67.25 & 68.04 \\
\hline
\end{tabular}
\caption{Performance (test accuracy(\%)) comparison with baselines. The three numbers in the next row of each dataset indicate the amount of labeled data; UDA(TextCNN) means applying UDA to TextCNN; TextCNN(fully) is a supervised method that uses the full dataset for training; TextCNN(KD) means distilling the knowledge of the inspirer into TextCNN; MixText uses the \{7,9,12\} layer for TMix; UDA(ALBERT) means applying UDA to ALBERT; UDA+KD(DistilBERT$_\text{6}$) means performing the KD method of 6 layers  DistilBERT to get a smaller and lighter model from BERT trained by UDA; UDA+KD(TinyBERT$_\text{4}$) means performing the KD method of 4 layers TinyBERT to get a smaller and lighter model from BERT trained by UDA; FLiText is our proposed method.}\label{tab:accents}
\end{table*}
\end{small}
\begin{table}
\begin{small}
\centering
\renewcommand{\arraystretch}{1.2}
\setlength\tabcolsep{3pt}
\begin{tabular}{l|c|c|c}
\hline
\textbf{Model} & \textbf{\#SpeedUp} & \textbf{\#Params} & \textbf{\#FLOPs}\\
\hline
UDA(BERT) & 1.0$\times$ & 110.08M & 22.5B \\
MixText & 1.0$\times$ & 85.7M & 24.3B \\
UDA(ALBERT) & 1.2$\times$ & 12.2M & 20.7B \\
UDA+KD(DistilBERT$_\text{6}$) & 2.3$\times$ & 52.7M & 11.3B \\
UDA+KD(TinyBERT$_\text{4}$) & 10.2$\times$ & 14.4M & 1.2B \\
UDA+KD(TinyBERT$_\text{6}$) & 2.1$\times$ & 67.5M & 11.3B \\
FLiText & 67.2$\times$ & 9.6M & 0.5B \\
\hline
\end{tabular}
\caption{Inference speed with Intel(R) Xeon(R) Platinum 8163 CPU @2.50GHz.}
\end{small}
\end{table}
We evaluate FLiText and baselines under different numbers of labeled data. The amount of labeled data is 20, 500, and 2500 respectively for IMDb, 500, 1000, and 2500 respectively for Yelp-5 and Yahoo. All of the amounts of unlabeled data is 70000. The experimental results are shown in Table 2.
\paragraph{Compared with the supervised learning method.}The results of FLiText and TextCNN(fully) in Table 2 show that, with only 500 labeled data, FLiText greatly exceeds the fully supervised method on the performance by 6.59\%, 3.94\%, and 3.22\% for each dataset respectively. Also, as the size of the labeled data increases to 2500, the performance is further improved to be 7.06\%, 6.28\%, and 6.18\%. This shows that FLiText is an effective SSL method for lightweight models.
\paragraph{Comparison with existing SOTA SSL methods on TextCNN.}Since the TMix method proposed by MixText cannot be directly applied to TextCNN, we apply the UDA framework to TextCNN. Among the results of FLiText, UDA(TextCNN) and TextCNN, there are two major findings. Firstly, when using 500 labeled data, FLiText achieves an accuracy improvement of 11.8\%, 18.26\%, and 9.78\%, compared to UDA(TextCNN) on the three datasets respectively. Secondly, in contrast, when using 2500 labeled data (5 times of FLiText), the accuracy of UDA(TextCNN) on the IMDb and YELP-5 is 3.16\% and 6\% lower than TextCNN(fully) respectively. This shows that, due to the limited feature extraction capabilities of the model, the application of UDA to TextCNN does not work. These two results show that FLiText is a SOTA semi-supervised text classification framework for lightweight models.
\paragraph{Comparison with existing SOTA SSL methods.}In this part, we compare the performance of FLiText, UDA, and MixText on the three datasets. Three conclusions are drawn from Table 2. Firstly, FLiText performs better on the IMDb and Yelp-5 datasets. For example, with 500 labeled data, the accuracy of FLiText on Yelp-5 is 1.53\% and 3.72\% higher than UDA and MixText, respectively. Secondly, on the IMDb dataset, as the number of labeled data decreases, FLiText has a more obvious advantage in performance compared with the other two methods. The same phenomenon can be observed on the Yelp-5 dataset. This shows that FLiText has a stronger ability to capture text features in scenarios with a few labeled data. Third, we also find that due to the relatively high difficulty for text classification of the Yahoo dataset with multiple 10 categories, the performance of FLiText is 1\% to 2\% lower than UDA or MixText under the three different volumes of labeled data. Overall, the performance of FLiText surpasses or approaches that of the SOTA frameworks on the semi-supervised text classification benchmarks, while the model obtained by FLiText is lighter (the scale of the parameters is only one-thousandth of UDA or MixText), and faster (the inference speed is 67 times faster than UDA or MixText). As a result, FLiText is a very practical framework, suitable for many actual industrial scenarios, especially in resource-limited scenarios or large-scale online systems, such as e-commerce search and real-time recommendation systems.
\paragraph{Comparison with the lightweight BERT.}
As shown in Table 2, the lightweight BERT (ALBERT) does not perform well under the framework of UDA, and is worse than that of FLiText. For example, when using 500 labeled data, FLiText achieves accuracy improvement of 1.7\%, 6.98\% and 2.95\%, compared to UDA(ALBERT) on the three datasets respectively. Moreover, even the base version of ALBERT has the same inference speed as BERT, which is 52 times of our method, as shown in Table 3.

\paragraph{Comparison with the KD method for BERT.}In the experiment, we also performed the KD methods of DistilBERT$_\text{6}$, TinyBERT$_\text{4}$, and TinyBERT$_\text{6}$ to get smaller and lighter models from BERT trained by UDA, to compare with FLiText. As shown in Table 2, the performance of ``UDA+KD(DistilBERT$_\text{6}$)'' is worse than FLiText under almost all experimental conditions, where the accuracy of the former is at least 0.5\% lower than that of the latter. The same conclusion can also be seen in the comparison with TinyBERT$_\text{4}$, which is the fastest variant of BERT in our experiment as shown in Table 3. Compared with TinyBERT$_\text{6}$, FLiText performs much better on the IMDb and Yelp-5 datasets. Though the performance of FLiText is about 1\% lower than TinyBERT$_\text{6}$ on the dataset of Yahoo, it is $32\times$ faster and $7\times$ smaller than TinyBERT$_\text{6}$, which is a valuable trade-off in the situations with low resources.

\paragraph{Comparison of the efficiency.}
From the results of Table 3, FLiText is $11.5\times$ smaller and $67.2\times$ faster than that of UDA(BERT), and it performs as well as UDA(BERT) on the datasets of IMDb and Yelp-5 with only 2.7\% FLOPs. Compared with ``UDA+KD(TinyBERT4)'', which is the smallest variant of BERT in Table 2 and Table 3, FLiText is $2\times$ smaller and $6.7\times$ faster, and achieves accuracy improvement of about from 1.5\% to 3\% on the three datasets with 46.2\% FLOPs. In terms of computational complexity, ~\citep{transformer} shows that Multi-Head Self-Attention requires $\mathcal{O}(n^2d+nd^2)$ operations while 1D-CNN requires $\mathcal{O}(k*n*d)$ operations, where n is the sequence length, d is the representation dimension, k is the kernel size of convolutions. Therefore, the computational complexity of BERT is $\mathcal{O}(L*(n^2d+nd^2))\approx\mathcal{O}(L*n*n*d)$, where L is the number of Transformer blocks, and the complexity of TextCNN is $\mathcal{O}(N*k*n*d)$, where N is the number of CNN kernels. Considering that $N*k << L*n$ in our situation, so the computational complexity of FLiText is much smaller than that of UDA(BERT).

\section{Ablation Study}
\subsection{Different Combinations of Transformer Layers and CNN Filters}
We choose different transformer layers and filters for multiple combinations and the results are shown in Table 4. We use \{Transformer layer\}-\{filter size\} to indicate which Transformer layers and CNN filters are selected. For example, \{0,1,2\}-\{2,3,5\} means that the first, second, and third layers of BERT are combined with the size of 2, 3, and 5 respectively. Three conclusions are drawn from Table 4. Firstly, a combination of a higher-level Transformer layer and a filter with a larger size achieves better performance. For example, with the combination of \{0,1,2\}-\{2,3,5\}, FLiText just achieves accuracy of 63.4\%; with \{6,7,8\}-\{5,7,9\}, the accuracy is increased from 63.4\% to 64.44\%; and with \{9,10,11\}-\{7,9,11\}, the accuracy is yet again increased to 64.53\%. Secondly, our hypothesis that filters with small window sizes focus on simple features, and as the filters become larger, the more advanced features can be captured. We performs inverted combinations, such as \{9,10,11\}-\{2,3,5\} and \{0,1,2\}-\{7,9,11\}, the accuracy is 63.15\% and 63.53\%, ~respectively, which is lower than the combination of \{0,1,2\}-\{2,3,5\}. Finally, FLiText achieves the best accuracy of 65.07\% with the combinations of \{5,7,9,11\}-\{2,3,9,11\}. This combination features BERT's middle-level and high-level transformer, as well as medium and large filters. Most of the grammatical and semantic information is transferred from BERT to TextCNN.
\begin{table}
\centering
\begin{tabular}{ccc}
\hline
\textbf{Transformer} & \textbf{TextCNN} &\textbf{Accuracy} \\
\hline
0,1,2 & 2,3,5 & 63.53 \\
3,4,5 & 3,5,7 & 64.05 \\
6,7,8 & 5,7,9 & 64.44 \\
9,10,11 & 7,9,11 & 64.53 \\
9,10,11 & 2,3,5 & 63.15 \\
0,1,2 & 7,9,11 & 63.4 \\
5,7,9,11 & 2,3,9,11 & 65.08 \\
1,3,5,7,9,11 & 2,3,5,7,9,11 & 64.51 \\
\hline
\end{tabular}
\caption{Accuracy on Yahoo with 500 labeled data and 70000 unlabeled data with different combinations of transformer layer and filters.}\label{tab:accents}
\end{table}
\subsection{Remove Different Parts of FLiText}
\begin{table}
\centering
\begin{tabular}{cc}
\hline
\textbf{Model} & \textbf{Accuracy} \\
\hline
FLiText & 67.25 \\
-feature distillation & 65.75 \\
-consistency regulation & 66.01 \\
-output distillation & 35.28 \\
\hline
\end{tabular}
\caption{Accuracy on Yahoo with 1000 labeled data and 70000 unlabeled data with remove different part.}
\end{table}

In order to verify the performance of each part of FLiText, we remove each component and show the results in Table 5. The removal of output-based distillation results in the worst performance degradation of the performance, which manifests that the target network mainly learns from the output-based knowledge The performance decreased 1.24\% after removing consistency regularization, which indicates that consistency regularization constraints can boost the performance in the KD framework. After removing the feature distillation, the performance dropped from 67.25\% to 65.75\%. This shows that feature distillation can help FLiText transfer more knowledge from the inspirer network to the target network on the basis of output-based distillation. 

\subsection{Consistency Regularization Effect}
We add a consistency regularization framework on the basis of KD to verify the performance of the former on the latter. 
\begin{table}
\centering
\begin{tabular}{cccc}
\hline
\textbf{Model} & \textbf{Yelp-5} & \textbf{Yahoo}\\
\hline
$TextCNN_{(KD)}$ & 57.98 & 65.81 \\
$TextCNN_{(KD+CR)}$ & 58.64 & 66.09 \\
\hline
\end{tabular}
\caption{Accuracy on Yelp-5 and Yahoo with 1000 labeled data and 70000 unlabeled data. $TextCNN_{(KD+CR)}$ represents adding consistency regularization on $TextCNN_{(KD)}$}
\end{table}
The results are shown in Table 6. We observe that after the introduction of the consistency regularization, the accuracy of $TextCNN_{(KD+CR)}$ is increased by 0.22\%, 0.66\% and 0.28\% on the three datasets respectively, compared to the $TextCNN_{(KD)}$.  
In our opinion, the improvement of performance brought by the consistency regularization is task-independent and can be used as a supplement to KD, so as to guide the student model to achieve better local smoothness.

\subsection{Nonlinear Activation Function Effect}
We find that adding a nonlinear transformation to the feature projection has a distinct impact on the performance of the model. In order to verify this impact, we examine the effect of using Relu, Tanh and avoid nonlinear transformation ($\varnothing$). The results are shown in Table 7. It can be seen that when the nonlinear transformation is removed, FLiText only achieves an accuracy of 63.37\%. Using Relu or Tanh offers a 1.71\% or 1.07\% boost in performance respectively.

\begin{table}
\centering
\begin{tabular}{cc}
\hline
\textbf{Model} & \textbf{Accuracy} \\
\hline
$\varnothing$ & 63.37 \\
$Relu$ & 65.08 \\
$Tanh$ & 64.44 \\
\hline
\end{tabular}
\caption{Accuracy on Yahoo with 500 labeled data and 70000 unlabeled data. $\varnothing$ is non-linear transformation.}
\end{table}


\section{Conclusion}
SSL has made great progress, but its rapid development is accompanied by increasingly complex algorithms and a sharp increase in the amount of computation, which is undoubtedly a bottleneck to the actual use of these algorithms in the industry. Therefore we introduce FLiText, a light and fast SSL framework for text classification with a convolution network. We show that FLiText achieves new SOTA results on multiple benchmark datasets on a lightweight model. Moreover, FLiTex achieves a close or even better performance compared to the previous SOTA SSL methods, while maintains a lightweight architecture with only one-thousandth of the parameters and a speed boost of more than 50 times. FLiText provides an effective way to deploy semi-supervised algorithms on resource-limited devices and industrial applications. In future research, we plan to apply FLiText to a wider range of NLP tasks, such as relation extraction and machine translation.

\bibliography{anthology,custom}
\bibliographystyle{acl_natbib}




\end{document}